\documentclass[sigconf]{acmart}
\usepackage{algorithm,balance,pbox}
\usepackage{algorithmicx,xcolor}
\usepackage{algpseudocode}
\usepackage{verbatim}
\usepackage{geometry}  % margin setting
\usepackage{soul,tabu}
\usepackage{booktabs} % For formal tables
\usepackage{xcolor}
\usepackage{textcomp,lscape,bm}
\usepackage{amssymb}
\usepackage{subfigure}
\usepackage{epsfig}
\usepackage{multirow}
\usepackage{colortbl}
\colorlet{shadecolor}{gray!20}
\usepackage{pifont}
\usepackage[utf8]{inputenc}
\usepackage{setspace}
\usepackage{marvosym}
\usepackage{ifsym,tabularx}
\usepackage{fontawesome}
\usepackage{amsmath,amssymb,amsfonts}
\usepackage{makecell}
\usepackage{adjustbox}
\usepackage{sidecap}
\usepackage{enumitem}
\usepackage{textcomp}
\usepackage{threeparttable}
\usepackage{fancyhdr}
\usepackage{todonotes}
\usepackage{color}
\usepackage[stable]{footmisc}
\usepackage{graphicx}
\usepackage{microtype}

\setlist[itemize]{leftmargin=*}
\usepackage{color}
\newcommand{\m}{{\sf {Hydra}}}
\newcommand{\guoqing}[1]{\textcolor{orange}{\bf [GZ: #1]}}

\newcommand{\kai}[1]{\textcolor{red}{\bf [kai: #1]}}
\newcolumntype{b}{>{\hsize=2.0\hsize}X}
\copyrightyear{2020}
\acmYear{2020}
\setcopyright{acmcopyright}\acmConference[SIGIR '20]{Proceedings of the 43rd International ACM SIGIR Conference on Research and Development in Information Retrieval}{July 25--30, 2020}{Virtual Event, China}
\acmBooktitle{Proceedings of the 43rd International ACM SIGIR Conference on Research and Development in Information Retrieval (SIGIR '20), July 25--30, 2020, Virtual Event, China}
\acmPrice{15.00}
\acmDOI{10.1145/3397271.3401121}
\acmISBN{978-1-4503-8016-4/20/07}
\renewcommand{\shortauthors}{K. Shu et al.}
\begin{document}

\title[Learning with Weak Supervision for Email Intent Detection]{Learning with Weak Supervision for Email Intent Detection}

% \author{Kai Shu, Subhabrata Mukherjee, Guoqing Zheng, Ahmed Hassan Awadallah, Milad Shokouhi and Susan Dumais}
\author{Kai Shu}
\authornote{Research was conducted while interning at Microsoft Research AI.}
\affiliation{
\institution{Arizona State University}
\city{Tempe}\state{AZ}
}
\email{kai.shu@asu.edu}

\author{Subhabrata Mukherjee}
\authornote{Equal contributions.}
\affiliation{
\institution{Microsoft Research AI}
\city{Redmond}\state{WA}
}
\email{submukhe@microsoft.com}

\author{Guoqing Zheng}
\authornotemark[2]
\affiliation{
\institution{Microsoft Research AI}
\city{Redmond}\state{WA}
}
\email{zheng@microsoft.com}

\author{Ahmed Hassan Awadallah}
\affiliation{
\institution{Microsoft Research AI}
\city{Redmond}\state{WA}
}
\email{hassanam@microsoft.com}

\author{Milad Shokouhi}
\affiliation{
\institution{Microsoft}
\city{Bellevue}\state{WA}
}
\email{milads@microsoft.com}

\author{Susan Dumais}
\affiliation{
\institution{Microsoft Research AI}
\city{Redmond}\state{WA}
}
\email{sdumais@microsoft.com}

% \settopmatter{printacmref=false}

\begin{abstract}

Email remains one of the most frequently used means of online communication. People spend significant amount of time every day on emails to exchange information, manage tasks and schedule events. Previous work has studied different ways for improving email productivity by prioritizing emails, suggesting automatic replies or identifying intents to recommend appropriate actions. The problem has been mostly posed as a supervised learning problem where models of different complexities were proposed to classify an email message into a predefined taxonomy of intents or classes. The need for labeled data has always been one of the largest bottlenecks in training supervised models. This is especially the case for many real-world tasks, such as email intent classification, where large scale annotated examples are either hard to acquire or unavailable due to privacy or data access constraints. Email users often take actions in response to intents expressed in an email (e.g., setting up a meeting in response to an email with a scheduling request). Such actions can be inferred from user interaction logs. In this paper, we propose to leverage user actions as a source of weak supervision, in addition to a limited set of annotated examples, to detect intents in emails. We develop an end-to-end robust deep neural network model for email intent identification that leverages both clean annotated data and noisy weak supervision along with a self-paced learning mechanism. Extensive experiments on three different intent detection tasks show that our approach can effectively leverage the weakly supervised data to improve intent detection in emails.

\end{abstract}

% \keywords{Weak supervision, user interaction, intent classification}
%\keywords{Intent detection, weak supervision, communication understanding}

\maketitle

\section{Introduction}
%sm: why weak supervision
%\input{figure_example}
\begin{figure}
\includegraphics[width=\columnwidth]{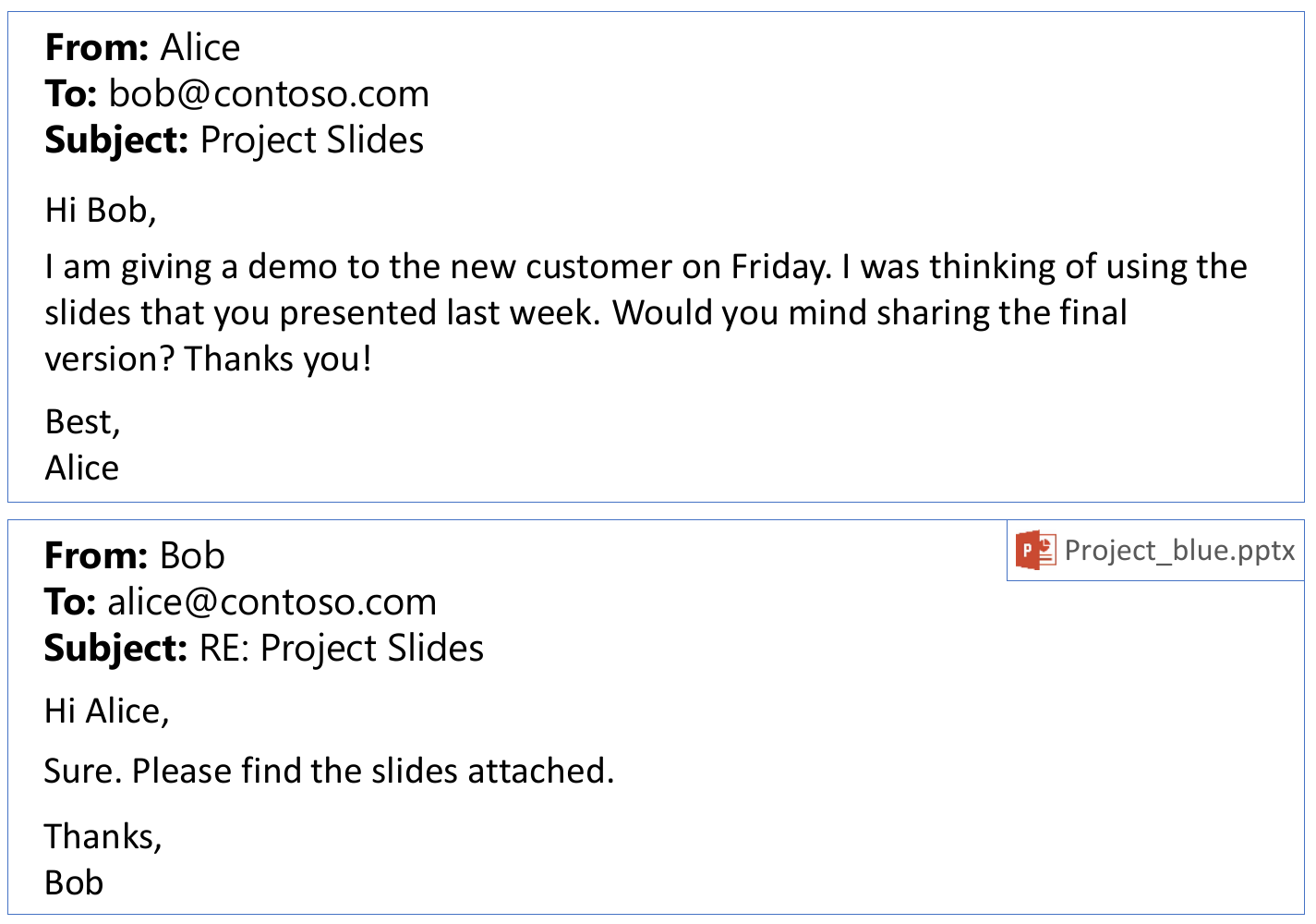}
 \vspace{-1.45\baselineskip}
	\caption{An illustration of user interaction for email intent classification. On the upper side, the sender Alice is \textit{requesting information} from the recipient Bob. On the lower side, the recipient Bob is replying Alice with an attachment -- an interaction that can be leveraged for weak supervision.}\label{fig:scenario}
        % \vspace{-0.2cm}
	    \vspace{-1.5\baselineskip}
%	    \vspace{-\baselineskip}
\end{figure}
%\begin{figure}[tbp!]
%	\centering
%	\subfigure[The sender has intent for requesting information.]{
%	\includegraphics[width=0.31\textwidth]{request_info.png}}\label{fig:request_info}
%	\subfigure[The recipient is replying with an attachment]{
%	\includegraphics[width=0.31\textwidth]{reply_with_attach.png}}\label{fig:reply_with_attach}
	%\vspace{-0.23cm}
%	\caption{An illustration of user interaction for email intent classification. On the upper side, the sender Alice is \textit{requesting information} from the recipient Bob. On the lower side, the recipient Bob is replying Alice with an attachment -- an interaction that can be leveraged for weak supervision.}\label{fig:scenario}
%	\vspace{-0.55cm}
%\end{figure}

Email has continued to be a major tool for communication and collaboration over the past decades. The volume of email messages exchanged daily has also continued to grow and is projected to reach 306.4 billion messages and 319.6 billion messages a day by the end of 2020 and 2021 respectively~\cite{Radicati2015}. In addition to the significant volume of messages, email is one of the top time consuming activities for information workers. Recent studies show that communicating with colleagues and customers takes up to 28\% of information workers' time, second only to role-specific tasks at 39\% ~\cite{McKinsey2012}. 

Such widespread use and significant amount of time spent on email have motivated researchers to study how people use email and how intelligent experiences could assist them to be more productive~\cite{Lampert:2010:DEC:1857999.1858140, Dabbish:2005:UEU:1054972.1055068,kannan2016smart}. One of the earliest works to characterize the main purpose email serves in work settings is that of Dabbish et al.~\cite{Dabbish:2005:UEU:1054972.1055068}. They conducted a survey of 124 participants to characterize different aspects of email usage. Based on this, they identified four distinct uses of email: task management, social communication, scheduling, and information exchange. More recent work~\cite{Wang2019ContextAwareII} conducted a large scale analysis of enterprise email identifying several use cases with many sub intents such as requesting an action, promising an action, updating a meeting, requesting information, social interaction, etc. Many other studies have focused on proposing methods for detecting intent of or suggesting actions in response to an email~\cite{Bennett:2005:DAE:1076034.1076140,Cohen04learningto,Lampert:2010:DEC:1857999.1858140,sappelli2016assessing}. Detecting intents in communications can integrate machine intelligence into email systems to build smart email clients that provide more value to email users. Several such applications have been studied including creating intelligent experiences that offer to assist users with scheduling a meeting~\cite{Horvitz:1999:PMU:302979.303030}, detecting action items~\cite{Bennett:2005:DAE:1076034.1076140}, automatically populating to-do lists~\cite{lampert2010detecting}, creating alerts for high-priority messages~\cite{Horvitz:1999:AA:2073796.2073831}, sharing documents~\cite{wang2019context}, and answering questions~\cite{yang2018characterizing}. 

Previous work posed email intent classification  as a supervised learning problem where human annotators were asked to annotate email messages given a predefined taxonomy and machine learning models were built to identify intents using the annotated dataset for training. Supervised models, especially those employing deep neural networks, rely on large scale annotated training data for learning. In many applications, manual annotation is either expensive and time-consuming to acquire, or infeasible due to privacy concerns for sensitive data. This is exactly the case for email data since its personal and private nature makes it hard to collect human annotations for. Even when annotations are collected, they are done on a limited amount of data that may not be sufficient or representative of the different domains.

Many application domains like recommendation, search, and email communication have rich user activities and interactions that can provide additional signals for learning~\cite{jiang2012social,agichtein2006improving,kooti2015evolution,ai2017characterizing}. For example, leveraging user interaction (e.g., clicks) for web search ranking has been extensively studied~\cite{agichtein2006improving}. Most email clients also allow users to manage their calendars, task lists, etc. Users interact with these information in different ways including responding, forwarding, flagging emails, setting up appointments, etc. Many of these user actions are directly correlated with the intent of the email. For example, many scheduling intents could be correlated with taking an action on the user's calendar such as creating or updating a calendar item. These actions may correlate to a certain intent but are also noisy in nature. Refer to Figure~\ref{fig:scenario} for an example. Consider the scenario where we want to detect emails where the sender is requesting a document from the recipient. In the absence of enough annotated examples, we may heuristically label all emails that received a response email {\em with attachments} as positive examples. However, there would be a lot of false positives since users may send attachments even without being requested for. Similarly, there will also be many false negatives, as users may not always send attachments even when requested to do so. 
In this work, we pose the email intent detection as a weakly supervised learning problem. We assume, as is usually the case in practice, that we have access to a small amount of annotated training examples for any given intent and a lot of weakly labeled noisy instances constructed from interaction-based heuristics (such as receiving a reply with an attachment). Note that such interaction signals typically cannot be used as features since they are only available after the user has processed and interacted with the message (e.g.,  predicting that one should schedule a meeting in response to a message after she has already done so is useless). Also, note that these weak supervision signals can be used by a model for training without requiring any human inspection alleviating a lot of privacy concerns. We propose an end-to-end trainable framework with deep neural networks to model these two sources of supervision simultaneously.

In summary, this paper makes the following contributions:
% \ms{can we really claim the first one as a novel contribution?}
% \noindent{\bf Contributions:} 
\begin{itemize}
    \item {\bf Application.} We show that weak supervision from user interaction is effective in the presence of limited amount of annotated data for the task of email intent identification. This pushes the frontier on weak supervision on email-related tasks where the focus has traditionally been on training fully supervised models. 
    \item {\bf Model.} We propose a unified framework {\m} to leverage cleanly annotated examples and weakly labeled ones {\em jointly} by embedding them in a shared representation space. We incorporate ideas from prior work on label correction to better handle noisy labels.%\guoqing{This leaves the impression that label correction is optional in the modeling. Since we removed Hydra-base, should we rephrase here? SM. Done.}
    \item {\bf Learning.} We propose a learning mechanism for {\m} based on prior works in curriculum learning and self-paced learning~\cite{kumar2010selfpaced}
    % \todo{GZ: these two appear here first. Either mention them earlier or provide citations}
    to judiciously select informative weak instances to learn from. 
    % \ms{Again, is this the first joint-approach? If not, what's the difference compared to state-of-the-art?}
    \item {\bf Experiments.} We conduct extensive experiments on real-world datasets to demonstrate the effectiveness of the proposed approach -- obtaining an accuracy improvement of $3\%$ to $12\%$ on average over state-of-the-art methods for different settings.
\end{itemize}

\section{Email Intent Detection}\label{sec:email}

In this section, we first formulate the weakly supervised learning problem of intent detection for email communication. Thereafter we describe in details the task, dataset, and how weak supervision can be leveraged from user interactions to help in the task.
%We choose the email communication domain which is rich in user activities and provide the necessary user interaction signals for weak supervision. %We will describe the annotation of clean labels, and the labeling functions for obtaining weak labels from the user interactions. Note that our approach is not specific to emails and can be extended to any arbitrary domain. 

\subsection{Problem Statement}
Let $\mathcal{D}=\{x_i, y_i\}_{i=1}^{n}$ denote a set of $n$ email messages with manually annotated clean labels, with $\mathcal{X}=\{x_i\}_{i=1}^{n}$ denoting the messages and $\mathcal{Y}=\{y_i\}_{i=1}^{n}$ the corresponding clean labels. Each message $x_i=\{w_1^i,\cdots,w_{m_i}^i\}$ contains a sequence of $m_i$ words. In addition to the small set of labeled examples, there is a large set of unlabeled examples. Usually the size of the clean labeled set $n$ is much smaller than the unlabeled set due to labeling costs or privacy concerns for email data. %For the large amount unlabeled data, weak labels derived from user interactions can be defined. (More details in Section \ref{sec:get_weak}).
%Note that in practice, we could generate arbitrary large amounts of weak supervision data given the large volume of available email messages and user interaction signals. However, for the sake of this paper, the volume of weak supervision data will be restricted by the size of the publicly avaialble email collection that we are using. More details about that will be discussed in Section~\ref{sec:experiments}.
For the widely available unlabeled samples, weak labels can be obtained based on \textit{user interactions} with emails (as illustrated in the example for Figure~\ref{fig:scenario} and more details in Section \ref{sec:get_weak}). Denote the weakly labeled set by
$\Tilde{\mathcal{D}}=\{\tilde{x}_j, \Tilde{y}_j\}_{j=1}^{N}$  where $\tilde{\mathcal{X}}=\{\tilde{x}_j\}_{j=1}^{N}$ denotes the set of $N$ unlabeled messages and  $\tilde{\mathcal{Y}}=\{\Tilde{y}_j\}_{j=1}^N$ denotes the set of weak labels derived from user interactions.
%$n<<N$. 
% \ms{Why are we switching between $\tilde{\mathcal{X}}$ and $\tilde{x}$ here? }
% where $\Tilde{y}^{k}_j$ is the weak label for instance $x^k_j\in \mathcal{X}^k$. 
%sm: This paragraph just repeats the problem statement below.
%Now our goal in this paper is to build a classifier $f:\mathcal{X}\rightarrow \mathcal{Y}$ in the presence of both a small clean set $\mathcal{D}$ and a larger weakly labeled set $\Tilde{\mathcal{D}}$ with its weak labels derived from user interactions. 
%In practice, it's always possible to obtain a small amount of task-specific labeled training data, and a large volume of unlabeled samples can be "weakly" labeled by the weak labeling function to serve as a weakly labeled set. 
%${\mathcal{X}}^k\ll \tilde{\mathcal{X}}^k, k=1,\cdots,K$. 
We now formally define our problem as:

%Typically, the size of clean set is much smaller compared to the weak supervision set ($m\ll M^k, k=1,\cdots,K$) due the scarcity of the expert labels, and the time-consuming as well as labor-intensive of human annotation. Formally, we can represent the problem as \textit{Weak-supervised Intent Detection}:

%\todo{GZ: Do we really want to emphasize that we are looking at multiple weak labeling functions here?}
%%\vspace{0.1in}
\begin{center}
\fbox{\parbox[c]{.9\linewidth}{\textbf{Problem Statement:}
Given a small manually annotated data $\mathcal{D}$ and a large set of weakly labeled data $\Tilde{\mathcal{D}}$ with weak labels derived from user interactions, learn an intent classifier 
$f: \mathcal{X}\rightarrow \mathcal{Y}$ which generalizes well onto unseen samples.
}}
\end{center}
  
\subsection{Dataset}

To better understand how real-world enterprise email data exhibits user intents, we leverage the Avocado\footnote{Avocado is a more appropriate test bed than Enron~\cite{klimt2004enron} since it contains additional meta-data and information beyond email such as calendar, tasks, etc.} %and it entered the public domain via the cooperation and consent of the legal owner of the corpus.} 
dataset~\cite{oard2015avocado}, which contains an anonymized version of the Outlook mailbox for 279 employees with various meta information. The full Avocado corpus contains $938,035$ emails, $26,980$ meeting schedules, and $325,506$ attachments. We focus on multiple {\em intent detection} tasks on this data and accordingly devise weak labeling functions  from user interactions.
% \ms{We should check with Peter Bailey about publishing based on Avocado data. Last time I heard there were concerns, but maybe only related to publishing "content" from Avocado emails.}
% \aha{We have published with the Avocado data multiple times already. We can publish results based on teh data but we cannot include any content derieved from it in the publication even as examples}

\begin{table*}[h]
\centering \caption{Examples of different intent types in enterprise emails with weak labeling rules derived from user interactions.}
\vspace*{-0.3cm}
\begin{tabular}{lll}
\toprule
 Intent & Example & Weak Supervision Rule\\
\midrule
Request Information (\textbf{RI}) & Please forward me the final version for the slides & reply\_with\_attachment\\
%\midrule
Schedule Meeting (\textbf{SM}) &  Let's meet on Monday to discuss these accounts & confirmed\_schedule\\
%\midrule
Promise Action (\textbf{PA}) & Once we have made reservations, we will let you know& urgency\_reply\\
\bottomrule
\end{tabular} \label{tab:intent}
\vspace*{-0.3cm}
\end{table*}

\noindent{\bf Intent types:} Prior works~\cite{dabbish2005understanding} have categorized email intents into four major categories: information exchange, task management, scheduling and planning, and social communications. Each category can have multiple fine-grained intents~\cite{wang2019context}. For instance, in the case of information exchange,  \textit{requesting information} is a common intent that indicates the sender is requesting information that can be potentially responded to by sharing a document. \textit{Schedule meeting} refers to the sender's intention to organize an event such as a physical meeting or a phone call, which belongs to the broader intent of scheduling and planning. In the case of task management intent, \textit{promise action} is an intent that indicates the sender is committing to complete a future action. In this work, we focus on these three intents -- request information, schedule meeting, and promise action (denoted by RI, SM, and PA, respectively). For instance, identifying the intent of requesting information allows an intelligent assistant system to automatically suggest files to share with the requester. This can result in improving the overall user experience and also user productivity. Table~\ref{tab:intent} shows some examples.

\subsection{Deriving Weak Labels from User Interactions}
\label{sec:get_weak}

With human annotations being hard to collect, if not impossible to obtain, in large scale, it is often cheaper and more beneficial to leverage weak supervision to build supervised models, particularly those based on deep neural networks. For email data, such weak supervision can be derived from user interactions. In this subsection, we discuss details to automatically obtain such weak labels from user interactions by using labeling functions and performing human evaluations to assess the quality of the weak labels.
\subsubsection{Weak Labeling Functions from User Interactions}\label{sec:weak}

For each of the aforementioned intent types, we define the weak labeling functions as follows:

{\em Request Information (RI):} We observe that the action of replying with attachments may potentially indicate the email it replies to has the intent of RI. %On one hand, from sender's perspective, a request information intent is to information from the receivers. 
For example, the email \emph{``Please forward me the final version for the slides''} is asking the recipient(s) to send a file back to the sender.  Now, if a user replies with an email \emph{``Please find the paper draft as attached''} along with an attachment, then the replied-to email is likely to contain the RI intent. However, this rule will have false positives since a user may reply with attachments even without being asked for. Additionally, messages with an RI intent may not receive a reply with an attachment or even any reply. Formally, the {\em weak} labeling function is:

%\vspace{0.05in}
\begin{center}
\textit{\parbox[c]{.9\linewidth}{\textbf{reply\_with\_attachment:}
If an email $a$ is replying to another email $b$ with an attachment, then email $b$ is weakly-labeled with the RI intent.
}}
\end{center}

%\vspace{0.05in}

Note that we ignore the trivial attachments that are not likely to contain information related to RI intent (e.g. contact information, signatures, images, etc.).

{\em Schedule Meeting (SM):} Since we have access to not only user emails but also their calendars, we explore the temporal footprints of the scheduled meetings including the subject line of the meeting, time, location and attendees. However, the emails that propose meeting requests are not directly associated with the schedule information. Therefore, we take the subject lines of the schedules as a query and search the emails that contain similar subject lines. This reveals the confirmation emails sent after someone accepted the meeting request. We temporally order the sent email together with the confirmation, and treat the earlier one as having the SM intent. %Therefore, we consider all the emails that have similar subject lines  In addition, we also enforce the sending time of the email with SM intent should be earlier than the confirmed schedule. 
The corresponding weak labeling function is defined as:

%\vspace{0.05in}
\begin{center}
\textit{\parbox[c]{.9\linewidth}{\textbf{confirmed\_schedule:}
If an email $a$ has the same subject line with another email $b$ confirming the schedule where $a$ precedes $b$ in the timeline, then $a$ is weakly-labeled with the SM intent.
}}
\end{center}
%\vspace{0.05in}

% \aha{this is confusing. We need to more clearly explain it}
{\em Promise Action (PA):} Outlook allows users to maintain a task list of items they need to do later. Tasks can be added to the task list by either directly creating them or by \emph{flagging} emails that may contain future action items. 
%To-do list provides an easy way to devise a rule for this intent. With its absence from our dataset, we look at the {\em flags} set by the sender. 
The flags could be added by either the sender or the recipient of an email. We use this behavior as a proxy label for future action's intent. For example, given an email from the sender as ``Would you be able to present your work in the meeting next week?" with the urgency flag set and a response email as ``I can do this next week" -- we consider the latter email to have the PA intent. 
%can setup an {\em urgency flag} intended for the attention of the recipient. In case the recipient responds, we consider the response email to have the intent of promising an action. For example, given an email from the sender as ``Would you be able to present your work in the meeting next week?" with the urgency flag set and a response email as ``I can do this next week" -- we consider the latter email to have the PA intent.
% \ms{this does nto make much sense to me.} \textbf{TODO: add more explanations}. 
The corresponding weak labeling function is defined as:

%\vspace{0.05in}
\begin{center}
\textit{\parbox[c]{.9\linewidth}{\textbf{urgency\_reply:}
If an email $a$ has replied to an email $b$ which had a follow-up flag set, then $a$ is weakly-labeled with the PA intent. 
}}
\end{center}
%\vspace{0.05in}
% why actionable intents, what are the types of intents
It is worth mentioning that although these labeling functions are geared for intent detection for email communication, it is possible to devise similar or more sophisticated rules for different intents and even different domains. 
% \ms{I still find it this not very intuitive. Can we justify better or at least explain why this works?}

%Note that the above extracted rules are by no means the only rules for extracting weakly-labeled instances from email user interactions. In this paper, we focus on studying the mechanism to extract and utilize weak supervision to improve intent detection, and we leave other sources/types of weak supervision rules for future work.

\subsubsection{Quality of Weak Labeling Functions}
\label{subsec:quality}

%Emails with weak supervision labels my not necessary be accurate enough to help improve intent detection performance. Thus, 
We conduct a manual evaluation to measure the quality of the labels generated by the weak labeling functions. We apply the aforementioned labeling functions (see Figure~\ref{tab:intent}) to the Avocado corpus, and obtain the following weakly labeled {\em positive} instances: 8,100 emails for RI, 4,088 emails for SM, and 2,135 emails for PA. In addition, we treat the emails discarded by the weak labeling functions as {\em negative} instances. For each intent, we sample the same amount  of negative instances as the positive ones to construct a balanced dataset. Note that the total amount of weak labeled data depends on the overall size of the email collection, how prevalent an intent is and  the trigger rate of the labeling functions. In practice, developers may have access to a much larger pool of unannotated emails compared to the Avocado dataset containing mailboxes for only 279 users. This may enable generating even larger weakly labeled instances and potentially further improving the overall performance. We study the effects of the relative size of the clean and the weak data later in the experiments section.

To assess the quality of these weakly-labeled instances, we randomly select 100 emails from each of the positive and negative weakly-labeled sets that are sent for manual annotation. Each email is annotated by three people and their majority consensus is adopted as the final annotation label for the instance. Table~\ref{tab:confusion} shows the confusion matrix from manual annotation. The accuracy of the weak labeling functions for the three intents RI, SM, and PA are 0.675, 0.71, 0.63, respectively. We observe that the accuracy of the labeling functions, while far from perfect, is also significantly better than random (0.5) for binary classification. This indicates that the weak labeling functions carry a useful signal albeit it being noisy. 
% \ms{given the random sample, how large do we expect the error bars to be?}

\begin{table}%htbp!]
\centering \caption{Confusion matrix for human evaluation of weak labeling functions.}
\vspace{-1em}
%\begin{tabular}{|l|l|r|r|}
\begin{tabular}{llrr}
\toprule
Intent & Predictions & True Positive & True Negative \\
\midrule
%\hline
\multirow{2}{*}{\textbf{RI}}&  Positive & 36\% & 64\%  \\
%\cline{2-4}
& Negative & 1\% &99\% \\
\midrule
%\hline
%\hline
\multirow{2}{*}{\textbf{SM}}&  Positive & 46\% & 54\%  \\
%\cline{2-4}
& Negative & 4\% &96\% \\
\midrule%\hline
%\hline
\multirow{2}{*}{\textbf{PA}}&  Positive & 31\% & 69\%  \\
%\cline{2-4}
& Negative & 5\% &95\% \\
\bottomrule%hline
\end{tabular} \label{tab:confusion}
\vspace{-0.4cm}
\end{table}

\subsection{Incorporating Small Amount of Clean Labels}
\label{subsec:clean}
%\aha{we need to rewrite this as discussed emphasizing the small amounts of labels needed}

\begin{table}%tbp!]
\centering \caption{Email datasets with metadata and user interactions. Clean refers to manually annotated emails, whereas weak refers to the ones obtained leveraging user interactions.}
\begin{tabular}{lccccc}
\toprule
%   & \multicolumn{2}{c}{Enterprise Email} & \multicolumn{2}{c}{Digital Assistant}  \\
Email & Intent & \multicolumn{2}{c}{Train} & {Dev} & {Test}\\
& & Clean & Weak&  & \\
\midrule
Avocado & RI & 1,800 & 16,200 & 334 & 336\\
& SM & 908 & 8,176 & 1,008 & 1,010\\
& PA & 474 & 4,270 & 518 & 518\\\midrule
Enron & SM & 908 & -& 908 & 908\\\bottomrule
\end{tabular} \label{tab:data}
\vspace{-0.2cm}
\end{table}

% For our experiments, we tap into a data rich in user activities and interactions. To this end, we use the largest and newest publicly available email communication dataset Avocado~\cite{oard2015avocado}. it consists of emails and attachments taken from 279 accounts of a now defunct information technology company referred to as ``Avocado''. 
% We focus on intent classification as an application and experiment with the following intents (as described in Section~\ref{sec:weak}): (i) \textit{Request Information} (RI); (ii) \textit{Schedule Meeting} (SM); and (iii) \textit{Promised Action} (PA).

Training neural networks with only noisy labels is challenging since they have high capacity to fit and memorize the
noise~\cite{zhang2016understanding}. Hence, it is useful to also incorporate clean labeled data in the training process. However, as discussed earlier, it is hard to collect human annotated data at scale especially when we need to support many intents across thousands of domains/organizations due to resource and privacy constraints. To better reflect these constrains in our experimental setup, we collected manual annotations for different intents such that the annotated set constitutes at most 10\% of all the labeled samples (weakly as well as manually labeled). Note that the clean labeled and weakly labeled email instances have no overlap. 

We first selected $10,000$ Avocado email threads randomly and excluded them from the weak supervision data collection. Given the number of weakly labeled instances for each intent in Section~\ref{sec:get_weak}, we collected manual annotations on 10\% of the selected email threads. To this end, three annotators examined all the messages in each thread and annotated the first message with one or more of the intents as described above with majority votes deciding the final label.
%These intents are annotated by human experts who examine all the related email messages and determine if they contain relevant signals for the intent. %Please refer to~\cite{wang2019context} for the annotation details. Note that the task in~\cite{wang2019context} is to identify sentence-level intent, while in this paper we consider the intent in entire email message.\kai{annotation details needed}
%Each email instance is annotated by 3 annotators and a majority voting is used to determine the final label. 
%We also perform the inter-rater agreement test and the 
The Cohen's kappa coefficient for inter-annotator agreement for each task was greater than or equal to 0.61 indicating a substantial agreement among the annotators~\cite{cohen1960coefficient}.  %\ms{same agreement on all 3 tasks?}
%SM: The annotation statistics are from prior work, not this work
The statistics of the dataset used for our intent classification tasks are reported in Table~\ref{tab:data}. For each intent, there are more negative samples than positive ones. We down-sample the negative class to make the classes balanced for each task. Although the entire Avocado email collection is large, we have only a few manually annotated clean samples for each intent. This also motivates the idea of incorporating user interactions as weak supervision to build better predictive models than using the clean samples alone. Based on our dataset construction, clean labels account for 10\% and weak labels constitute 90\% of all the labeled samples (both clean and weak). Additionally, we report the performance of various methods in the extreme case -- where clean labels account for only 1\% of all the labeled samples obtained by further down-sampling the clean examples.
\footnote{We will make the code and data publicly available at \href{https://aka.ms/HydraWS}{https://aka.ms/HydraWS}, in accordance with the sharing policy for the Avocado dataset.}.

% \kai{refer to Wei's paper for more details; why this is a difficult task}

% \section{Joint Learning with Multiple Sources of Supervision}\label{sec:model}%\todo{Better name}

\section{Joint Learning with Clean and Weak Sources of Supervision}\label{sec:model}%\todo{Better name}
\begin{figure}[t]%tbp!]
	%\vspace{-0.25cm}
	\centering
	%\subfigure[{\m}-Base]{
	%\includegraphics[width=0.26\textwidth]{framework_base.png}}
	%\hspace{0.2cm}
		\subfigure[{\m} training]{
	\includegraphics[height=0.28\textwidth]{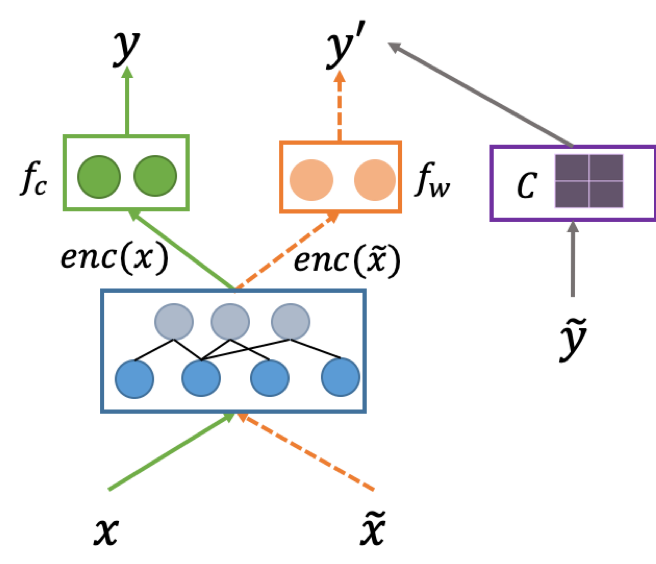}}
	%\hspace{0.33cm}
		\subfigure[{\m} inference]{
	\includegraphics[height=0.28\textwidth]{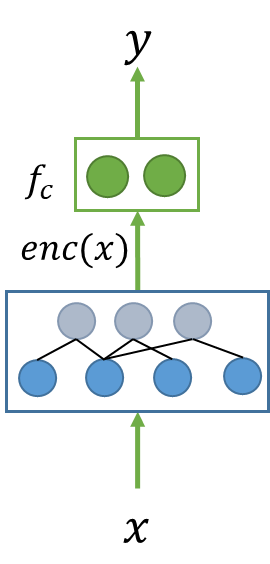}}
 	\vspace{-0.4cm}
	\caption{Proposed framework {\m} for learning with weak supervision from user interactions. 
	%(a) {\m}-Base leverages both clean ($x$) and weak examples ($\tilde{x}$) with a shared feature learning component (blue box) that generates corresponding encoded representations $enc(x)$ and $enc(\tilde{x})$, respectively. The encoded feature representations are used to learn functions $f_c$ and $f_w$ jointly optimized to model the clean ($y$) and weak labels ($\tilde{y}$), respectively; 
	(a) {\m} leverages a label correction component (purple box) to rectify weak labels ($\tilde{y} \rightarrow y'$) while learning; 
% 	\ms{would be good to color inside the box as purple too. It's unclear how $\tilde{y}$ and $y'$ interact.}
	(b): During inference, {\m} uses learned feature representation (blue box) and function $f_c$ to predict labels for (unseen) instances in test data.}\label{fig_framework}
	\vspace{-0.4cm}
\end{figure}

%We use labeling functions over {\em user interactions} to generate weakly labeled data automatically. This is used to augment the manually annotated examples and both of them are leveraged jointly for end-to-end learning. As an example of a labeling function, consider the scenario for an intent classification task where given an Email message, we intend to predict whether it is requesting for information. In this case, if a user replies with an attachment then the replied-to message is weakly labeled as positive for requesting for information. We defer the detailed discussion of designing weak labeling functions to the next section. 

Having defined the problem setting for intent detection with weakly supervised learning in the presence of a small set of cleanly labeled examples and a large set of weakly labeled ones, we now propose our approach to leverage these two sources of supervision jointly to learn an end-to-end model.
%In this section, we present the problem statement, and focus on developing algorithms to jointly leverage these weakly labeled instances and the manually annotated ones to obtain a robust model. 
%In addition, we will also explore label correction techniques to improve our model.

%the details of the proposed framework for \underline{W}eak-supervised \underline{I}ntent \underline{D}etection, named {\m}. It mainly consists of three components (See Figure~\ref{fig_framework}): (1) a label correction component; (2) a dual-source weak supervision modeling component; and (3) a weak supervision fusion component.

%Specifically first, the label correction component describes the modeling for correcting the weak labels by estimating a corruption matrix the encodes the correlations between clean labels and weak labels; next, the multi-task neural network component learns representations by jointly performs the tasks of predicting clean and weak labels simultaneously; finally, the weak supervision fusion component combine the label correction and multi-task learning component for indent detection.

% \subsection{Two-Stream Weak Supervision Modeling}
% \todo{GZ: maybe clarify the multiple sources here, or change to a more accurate name?}

\subsection{\m: Dual-source Supervised Learning}\label{sec:base}
%\aha{we need to explain/introduce the term Hydra here}
Multi-source learning has shown promising performance in various domains such as truth discovery~\cite{ge2013multi}, object detection~\cite{ouyang2014multi}, etc. In our email intent detection scenario, we have two distinct sources (dual-source) of supervision: clean labels coming from manual annotation and weak labels coming from heuristic labeling functions based on user interaction signals.

Our objective is to build a framework that leverages signals coming from both sources of supervision and learn an underlying common representation from the context. We develop a deep neural network where the lower layers of the network learn {\em common} feature representations of the input space (text of messages in our context), and the upper layers of the network {\em separately} model the mappings to each of the different sources of supervision. In this way, we are able to {\em jointly} leverage both the correlation and distinction between the clean and weak labels. Since the weak labels are obtained from labeling functions defined over user interactions, they contain complementary information to the clean labels annotated from message contents. We term this framework \m\footnote{{\m} is a multi-headed creature in the Greek legend. Our proposed framework leverages multiple sources of supervision jointly and therefore is named {\m}.}.

Recall $\mathcal{D}=\{x_i,y_i\}_{i=1}^n$ and $\Tilde{\mathcal{D}}=\{\Tilde{x}_j, \Tilde{y_j}\}_{j=1}^N$ to be the clean labeled data (based on manual annotation) and the weak labeled data (based on user interactions) respectively. 
%\aha{what do we mean by contextual here?}\kai{changed to content/text?}
Let $enc(x;\theta)$ to be an encoder that produces the content representation of an instance $x$ with parameters $\theta$. Note that this encoder is shared between instances from both the clean and the weak set. Let $f_c(enc(x); \gamma_c)$ and $f_w(enc(\Tilde{x}); \gamma_w)$ be the functions that map the content representation of the instances to their labels on the clean and weakly supervised data, respectively. Note that in contrast to the encoder with shared parameters $\theta$, the parameters $\gamma_c$ and $\gamma_w$ are different for the clean and weak sources respectively to capture their individual characteristics. The final objective for jointly optimizing the predictions from dual sources of supervision is given by:
\begin{equation}
\label{eq:0}
    \min_{\theta, \gamma_c, \gamma_w} \mathbb{E}_{(x,y)\in\mathcal{D}}\mathcal{L}(y,f_c(enc(\mathbf{x})))+\alpha\mathbb{E}_{(\mathbf{\Tilde{x}},\Tilde{y})\in\mathcal{\Tilde{D}}}\mathcal{L}(\Tilde{y},f_w(enc(\mathbf{\Tilde{x}})))
\end{equation}
where $\mathcal{L}$ denotes the loss function to minimize the prediction error of the model. $\alpha$ is a hyper-parameter that controls the relative importance of the loss functions computed over the data from clean and weak sources. 

\subsubsection{Weak Label Correction}\label{sec:hydra_glc}

% introduce glc
Labeling functions are heuristic and can generate false labels (refer to Section~\ref{subsec:quality} for an evaluation of labeling functions). An intuitive approach is to consider correcting these noisy labels before feeding them into the above model. Label correction methods have been previously studied for learning from noisy data sources~\cite{hendrycks2018using,ren2018learning}. 
% In Section~\ref{sec:glc}, we reviewed the Gold Loss Correction approach (GLC) from~\cite{hendrycks2018using}. In this, we want to estimate a label corruption matrix $\mathbf{C}\in\mathcal{R}^{L\times L}$ that encodes the correlations between clean and weak labels.
We now give a brief primer on existing work on label correction before discussing on how to integrate it into our framework.
%\kai{Since weak label correction approaches can be a building block that incorporates into our model, we first briefly introduce a representative prior work on learning with weak supervision on label correction.}

\noindent{\em Prior Work on Label Correction.}\label{sec:glc} Leveraging weak supervision for building effective supervised models has shown performance improvement in various tasks~\cite{hendrycks2018using,ren2018learning}. Hendrycks \textit{et al.} propose the idea of learning a {\em label corruption matrix} to estimate clean labels from the weak labels with the Gold Loss Correction approach (GLC)~\cite{hendrycks2018using}.
Given a set of instances $\mathcal{D}=\{x_i,y_i\}_{i=1}^n$ with manually annotated (clean) labels $y$ for $L$ categories, and a weak labeled set $\Tilde{\mathcal{D}}=\{\Tilde{x}_j, \Tilde{y_j}\}_{j=1}^N$, GLC aims to estimate a matrix $\mathbf{C}\in\mathcal{R}^{L\times L}$ to model the label corruption process. %To this end, we adopt the Gold Loss Correction approach (GLC)~\cite{hendrycks2018using}. 
Formally, we first train a classifier $f$ on the weakly labeled data $\Tilde{\mathcal{D}}=\{\Tilde{x}_j, \Tilde{y_j}\}_{j=1}^N$ as:  
% \ms{are we consistent in our notations here for weak labels vs. clean labels?}
\begin{equation*}
    f(\tilde{x}) = \hat{p}(\Tilde{y}|\tilde{x},\bm{\theta})
\end{equation*}
Let $\mathcal{X}_l$ be the subset of $x$ with label $y=l$. 
Assuming the conditional independence of $\Tilde{y}$ and $y$ given ${x}$, i.e., $p(\Tilde{y}|y,x) = p(\Tilde{y}|x)$, 
we then estimate the corruption matrix $\Tilde{\mathbf{C}}$ as follows,%$\Tilde{\mathbf{C}}_{lr} \approx p(\Tilde{y}=r|y=l)$.
% \begin{comment}
    \begin{align}
     \nonumber \mathbf{C}_{lr}& = \frac{1}{|\mathcal{X}_l|}\sum_{x\in\mathcal{X}_l}\hat{p}(\Tilde{y}=r|x)
     =  \frac{1}{|\mathcal{X}_l|}\sum_{x\in\mathcal{X}_l}\hat{p}(\Tilde{y}=r|y=l,x)\\
     & \approx p(\Tilde{y}=r|y=l)\label{eqn:c}
\end{align}
% \end{comment}
With the new estimated $\mathbf{C}$, they train a new classification model $f'(x)=\Tilde{p}(y|x,\theta)$ solving the following optimization problem:
\begin{equation}\label{eqn:glc}
    \min_{\theta} \mathbb{E}_{(x,y)\in\mathcal{D}}\mathcal{L}(y,f'(x))+\mathbb{E}_{(\tilde{x},\Tilde{y})\in\mathcal{\Tilde{D}}}\mathcal{L}(\Tilde{y},\mathcal{C}^\texttt{T}f'(\tilde{x}))
\end{equation}
where $\mathcal{L}$ is a differentiable loss function to measure the prediction error, such as the cross-entropy loss.

\begin{comment}
Essentially we do not apply correction on the clean instances in $\mathcal{D}$, while we correct the predicted labels for instances in $\mathcal{\Tilde{D}}$ by multiplying the corrupted matrix as in $\mathcal{\tilde{C}}^\texttt{T}f'(x)$. Finally, we use the resultant model $f'(x)$ as the label correction function for each instance $(x_j,\Tilde{y})\in\Tilde{\mathcal{D}}$ as,
\begin{equation}
    y'_j = f'(x)
\end{equation}
where $y'_j$ is the corrected label for $x_j$. Therefore, we can obtain a corrected weak supervision set $\mathcal{D}'=\{x_j, y'_j\}_{j=1}^N$.
\end{comment}

\begin{table}[tbp!]
\centering \caption{Notation Table.}
% \small
\begin{tabular}{ll}
\toprule
Notation & Meaning \\
\midrule
$\mathcal{D}$ & set of instances with clean labels \\
%\midrule
$\Tilde{\mathcal{D}}$& set of instances with weak labels \\
%\midrule
$\mathcal{D}'$ & set of instances with weak label correction\\
%\midrule
 $enc(\cdot)$& shared encoder for learning latent representations \\
 %\midrule
 $\mathbf{C}$& label corruption matrix \\
 %\midrule
 $f_c$ & function for predicting clean labels\\
  %\midrule
 $f_w$ & function for predicting weak labels\\
  %\midrule
 $f'$ & function for correcting weak labels\\
 $w$ & neural network model parameters\\
 $v$ & latent variable to select weak training samples\\
\bottomrule
\end{tabular} \label{tab:symbol}
%\vspace{-0.25cm}
\end{table}

% introduce Hydra after glc

\subsubsection{Integration with {\m}}
We use a similar idea and correct the weak labels for instances in $\mathcal{\Tilde{D}}$. Using Equation~\ref{eqn:glc}, we learn a label correction function $f'(\Tilde{x})$ that rectifies the weak labels coming from the labeling functions for each instance $\Tilde{x} \in \Tilde{\mathcal{D}}$. We now obtain a label corrected weak supervision set $\mathcal{D}'=\{\Tilde{x_j}, f'(\Tilde{x_j})\}_{j=1}^N$. Note that the label correction network reduces noise but the rectified labels could still be erroneous, and therefore considered as another source of weak supervision.
%Given a set of instances $\mathcal{D}=\{x_i,y_i\}_{i=1}^m$ with manually annotated (clean) labels $y$ for $L$ categories, and a weak labeled set $\Tilde{\mathcal{D}}=\{\Tilde{x}_j, \Tilde{y_j}\}_{j=1}^M$, we want to estimate a matrix $\mathbf{C}\in\mathcal{R}^{L\times L}$ to model the label corruption. To this end, we adopt the Gold Loss Correction approach (GLC)~\cite{hendrycks2018using}. 
%We have introduced how we can use label correction approach to obtain a higher quality weak labels, and how multi-source modeling can jointly optimize clean labels and weak supervision labels simultaneously. To further explore their mutual benefits, We further integrate these components together for intent prediction (see Figure~\ref{fig_framework}). 
In the new setting, we first feed the weakly labeled instances $\tilde{\mathcal{D}}$ from the labeling function into the label correction network to obtain the rectified instances $\mathcal{D}'$. These are used as an input to \m. Formally, the overall objective function of our final model {\m} is given by:
\begin{multline}\label{eqn:obj}
    min_{\theta, \gamma_c, \gamma_w} \mathbb{E}_{(x,y)\in\mathcal{D}}\mathcal{L}(y,f_c(enc(x)))+\\\alpha\mathbb{E}_{(\Tilde{x},f'(\Tilde{x}))\in\mathcal{D}'}\mathcal{L}(f'(\Tilde{x}),f_w(enc(\Tilde{x})))
\end{multline}

\subsection{Self-paced Learning for \m}
A simple training process is to consider all the weakly labeled samples jointly for learning. However, not all training samples are created equal. Some of the weak instances are noisier than others; whereas some are quite different in nature than the clean samples and therefore more difficult to learn from.  

This is similar to curriculum learning~\cite{bengio2009curriculum} where a training schedule is used to first learn from easy samples followed by difficult ones. The main challenge however is the distinction between easy and hard training samples. To alleviate this challenge, we can leverage the learned model to identify an easy set of samples given by a good fit in the model space similar to self-paced learning~\cite{kumar2010selfpaced}. %We leverage a similar idea for training \m. 

%We start with \m trained on a small clean set of samples and iteratively select instances from the pool of weakly labeled instances to inject into the training set. 
Consider $v(\Tilde{x}) \in \{0,1\}$ to be a latent variable for each weak instance $\Tilde{x}$ that dictates whether to consider it for training. Correspondingly, our objective function is modified as follows.
{\small
\begin{multline}\label{eqn:obj1}
    \min_{\theta, \gamma_c, \gamma_w, v \in \{0,1\}^N} \mathbb{E}_{(x,y)\in\mathcal{D}}\mathcal{L}(y,f_c(enc(x)))+\\ \alpha\mathbb{E}_{(\Tilde{x},f'(\Tilde{x}))\in\mathcal{D}'}[v(\Tilde{x}) \cdot \mathcal{L}(f'(\Tilde{x}),f_w(enc(\Tilde{x})))] - \lambda ||v||_1
\end{multline}
}
There are two distinct sets of parameters to learn corresponding to $w=\{\theta, \gamma_c, \gamma_w\}$ for the neural network parameters and latent variables $v$ for the training sample selection. To optimize the above equation, we employ alternate minimization. We first fix $v$ and estimate the model parameters $w$ using gradient descent. 

Next we fix $w$ and estimate $ v(\Tilde{x})$ for all $\Tilde{x} \in \Tilde{D}$. Partial derivative of Eqn.~\ref{eqn:obj1} with respect to $v(\Tilde{x})$ is given by $\alpha \mathcal{L}(f'(\Tilde{x}),f_w(enc(\Tilde{x}))) - \lambda $. The optimal solution for the equation is given by:
\[
    v(\Tilde{x})= 
\begin{cases}
    1,& \text{if } \mathcal{L}(f'(\Tilde{x}),f_w(enc(\Tilde{x}))) < \frac{\lambda}{\alpha}\\
    0,              & \text{otherwise}
\end{cases}
\]
%If $\mathcal{L}(f'(\Tilde{x}),f_w(enc(\Tilde{x}))) < \frac{\lambda}{\alpha}$, then the optimal value of the above equation is obtained at $v(\Tilde{x})=1$. Similarly, $v(\Tilde{x})=0$ gives the optimal solution for $\mathcal{L}(f'(\Tilde{x}),f_w(enc(\Tilde{x}))) > \frac{\lambda}{\alpha}$.

Here $\frac{\lambda}{\alpha}$ indicates whether an instance is easy to learn given by a small value of the corresponding loss function $\mathcal{L}(.)$. A high loss indicates a poor fit of the sample in the model space and therefore ignored during training. $\lambda$ as a hyper-parameter lets us control the injection of weak samples in the training set: a very low value admits few whereas a very high value admits all samples.

We initially train {\m} on only the clean data for a few epochs to trace the corresponding model space. Thereafter, we incorporate the weakly labeled samples gradually by increasing $\lambda \in \{0.1, 0.2 , \cdots \}$ till all samples are included in the training set. 

\subsection{\bf Training \m} 
\label{sec:training}
We adopt mini-batch gradient descent with Adadelta~\cite{zeiler2012adadelta} optimizer to learn the parameters. Adadelta is an adaptive method which divides the learning rate by an exponentially decaying average, and is less sensitive to the initial learning rate. %We choose Adadelta as the optimizer because it is a popular and effective methods for determining the learning abortively, which is widely used for training neural networks.
For ease of understanding, all the notations we use are summarized in Table~\ref{tab:symbol}.

We first train the GLC model to obtain the label corrected weak supervision set $\mathcal{D}'$. To this end, we train a classifier $f$ on weak supervision data $\tilde{\mathcal{D}}$ and estimate the label corruption matrix ${\mathbf{C}}$. Thereafter, we train a new classifier $f'$ with the corruption matrix on the weakly supervised data, and obtain the data with corrected weak labels $\mathcal{D}'$. 

Next we train {\m} for a few epochs on the clean data to have an initial estimate of $w$. Given $w$ and an initial value of $\lambda$, we compute loss for all the weak instances and include those with loss less than $\frac{\lambda}{\alpha}$ in the training set. This is followed by re-estimating $w$. We iterate over these steps and gradually increase $\lambda$ until all the samples are accounted for or the model stops improving. For inference, the label of an instance $x$ is predicted by $y=f_c(enc(x))$. 

%We train {\m} end-to-end in Lines~\ref{h_start} through Line~\ref{h_end}. We first sample each batch with equal number of $m$ instances from clean and weak sources, and feed them into {\m} network for training. For inference, the label of an instance $x$ is predicted by $y=f_c(enc(x))$. 

% \guoqing{this conflicts with the Algorithm 1} 

% \kai{add batch level description on the batch construction mechanism. may remove the GLC part}
%\vspace{-0.15cm}

\section{Experiments}\label{sec:experiments}

\begin{table*} [htbp!]
\centering \caption{Performance of the proposed approach compared to several baselines. Clean Ratio denotes the ratio of clean labels to all available labels (clean and weak) that is used to train the corresponding models. We show results for 10\% (\textbf{All})) and 1\% (Tiny) clean ratios. {\m} outperforms all the baselines in all settings.}  
%The columns for Clean (manually annotated) and Weak (from labeling functions) depict the model performance using only those subsets of the data; Clean+Weak is the simple union of both the sets where we consider weak labels to be as reliable as clean ones, whereas IWT assigns higher weights to the clean instances compared to the weak ones. PreWeak pre-trains the model on weak labels and fine tunes on clean ones. \guoqing{Hydra-Base to be replaced}}% \kai{Can we do stat. sig. test for the close ones?}
%\begin{tabularx}{0.98\linewidth}{*{8}{X}}
%\vspace{-0.25cm}
\begin{tabular}{cccccccccc}
\toprule%hline
Clean Ratio (Setting)& Intent & Encoder (enc)  & Clean & Weak  & Clean + Weak & Pre-Weak & IWT & GLC & {\m}  \\
% \midrule%\hline \hline
\cline{2-10}
 \multirow{6}{*}{\textbf{10\% (All)}} & \multirow{2}{*}{\textbf{RI}} & AvgEmb \cellcolor{shadecolor}& \cellcolor{shadecolor}0.649 &\cellcolor{shadecolor} 0.523 & \cellcolor{shadecolor}0.616 &\cellcolor{shadecolor}0.613 & \cellcolor{shadecolor}0.661 & \cellcolor{shadecolor}0.693 & \cellcolor{shadecolor} {\bf 0.726} \\%\textbf{0.716}
%\cline{2-8}
&& BiLSTM & 0.688 & 0.524 & 0.684 &0.717 & 0.711 & 0.717 & {\bf 0.804} %\textbf{0.754} 
\\
% \midrule%\hline\hline
\cline{2-10}
& \multirow{2}{*}{\textbf{SM}} & AvgEmb \cellcolor{shadecolor}& \cellcolor{shadecolor}0.650 & \cellcolor{shadecolor}0.624 & \cellcolor{shadecolor}0.691 &\cellcolor{shadecolor}0.676 & \cellcolor{shadecolor}0.713 & \cellcolor{shadecolor}0.694 & \cellcolor{shadecolor}{\bf 0.731}% \textbf{0.715} 
\\
%\cline{2-8}
&& BiLSTM &0.655  & 0.605 & 0.693 &0.702 & 0.705 & 0.697 & {\bf 0.714} %\textbf{0.713} 
\\
% \midrule%\hline\hline
\cline{2-10}
&\multirow{2}{*}{\textbf{PA}} & AvgEmb \cellcolor{shadecolor}& \cellcolor{shadecolor}0.641 & \cellcolor{shadecolor}0.628 & \cellcolor{shadecolor}0.633  &\cellcolor{shadecolor}0.637 & \cellcolor{shadecolor}0.625 & \cellcolor{shadecolor}0.647 & \cellcolor{shadecolor}{\bf 0.664}% \textbf{0.655}
\\
%\cline{2-8}
&& BiLSTM & 0.608 & 0.547 & 0.611 &0.631 & 0.616 & 0.635 & {\bf 0.660}%\textbf{0.650} 
\\\bottomrule
% \bottomrule
  \multirow{6}{*}{\textbf{1\% (Tiny)}} &\multirow{2}{*}{\textbf{RI}} & AvgEmb \cellcolor{shadecolor}& \cellcolor{shadecolor}0.560 & \cellcolor{shadecolor}0.523 & \cellcolor{shadecolor}0.529 &\cellcolor{shadecolor}0.542 &\cellcolor{shadecolor}0.563& \cellcolor{shadecolor}0.592 & \cellcolor{shadecolor}{\bf 0.664}% \textbf{0.657}
\\
%\cline{2-8}
&& BiLSTM & 0.539 & 0.524 & 0.560 &0.581 & 0.565 & 0.572 &  {\bf 0.622} %\textbf{0.633}
\\
% \midrule%\hline\hline
\cline{2-10}
&\multirow{2}{*}{\textbf{SM}} & AvgEmb \cellcolor{shadecolor}& \cellcolor{shadecolor}0.565 &\cellcolor{shadecolor} 0.624 &\cellcolor{shadecolor} 0.618 &\cellcolor{shadecolor}0.633 & \cellcolor{shadecolor}0.628 &\cellcolor{shadecolor} 0.620 & \cellcolor{shadecolor}{\bf 0.666}% \textbf{0.669} 
\\
%\cline{2-8}
&& BiLSTM &0.538  & 0.605 & {0.626} & 0.608 & 0.625 & { 0.617} &  {\bf 0.630} %\textbf{0.655}
\\
% \midrule%\hline\hline
\cline{2-10}
&\multirow{2}{*}{\textbf{PA}} & AvgEmb \cellcolor{shadecolor}& \cellcolor{shadecolor}0.584 & \cellcolor{shadecolor}0.628 & \cellcolor{shadecolor}0.633 &\cellcolor{shadecolor}0.616 & \cellcolor{shadecolor}0.622 & \cellcolor{shadecolor}0.613 & \cellcolor{shadecolor}{\bf 0.647}% \textbf{0.645}
\\
%\cline{2-8}
&& BiLSTM & 0.569 & 0.547 & 0.571 &0.573 & 0.577 & 0.587 & {\bf 0.626} %\textbf{0.611} 
\\
\bottomrule%\hline
\end{tabular} \label{tab:performance}
\vspace{-0.2cm}
\end{table*}

In this section, we present the experiments to evaluate the effectiveness of {\m}. 
\begin{comment}
Specifically, we aim to answer the following evaluation questions:
\begin{itemize}
    \item \textbf{EQ1}: Can {\m} improve classification performance by leveraging weak supervision from user interactions?
    \item \textbf{EQ2}: How effective are the clean and weak source of supervision for improving prediction performance? \guoqing{don't we only have one source of weak supervision?}\kai{I rephrase it}
    % \item \textbf{EQ3}: How robust is {\m} on leveraging weak supervision in the presence of noise?
\end{itemize}
\end{comment}

%To answer the first question, we compare the performance of {\m} with the state-of-the-art intent detection methods. We then investigate the effects of different sources of weak supervisions, respectively. Moreover, we demonstrate the performance of {\m} with different prediction functions.

%\aha{This seems pretty late since we discussed the data and its size already earlier. Can we move it up? One way to do that is to have section 3 structured as follows: 3.1: General description of the dataste (emails, calendars, etc.), 3.2: Weak labels and 3.3: Clean labeles. For 3.3, we should provide some information about how the data was labeled beyond just refering to the reference}

% \subsection{Datasets}\label{sec:data}

\subsection{Experimental Settings}\label{sec:setting}
\subsubsection{ Datasets.} We primarily perform experiments on the Avocado email collection. We perform experiments on three different tasks (intents):  request information, schedule meeting and promise action. Section~\ref{subsec:clean} discusses, in details, the data annotation process to obtain the clean labels and harnessing user interactions to obtain the weakly labeled instances. In addition to Avocado, we also perform an experiment to show the generalizability of our approach for transfer to another domain, namely the email collection for Enron (discussed in Section~\ref{subsec:dom_gen}). Table~\ref{tab:data} shows the dataset statistics. The datasets are balanced with equal number of positive and negative instances. Note that Avocado is the only public email collection with available user interaction logs available.

% evaluation metrics
\subsubsection{ Evaluation metric:} We pose our task as a binary classification problem for each of the intents. We use accuracy as the evaluation metric. %We randomly choose 80\% of the instances in the clean set for training, 10\% for validation, and remaining 10\% for testing. 
We report results on the test set with the model parameters picked with the best validation accuracy on the dev set (Table~\ref{tab:data} shows the data splits). All runs are repeated for 5 times and the average is reported. We compare different methods and techniques at different values of the {\em clean data ratio} defined as:
\begin{equation*}
    \text{clean ratio}=\frac{\text{\#clean labeled samples}}{\text{ \#clean labeled samples + \#weak labeled samples}}
\end{equation*}
Specifically, we report performance with two settings with different clean data ratios:
\begin{itemize}
\item \textbf{All}: The setting where all available clean labels are used. According to our dataset construction process in Section~\ref{subsec:clean} this corresponds to clean ratio = 10\%. 
\item \textbf{Tiny}: The extreme setting where clean labels are down-sampled to account for only 1\% of all data samples (clean and weak) to demonstrate the impact of weak supervision when extremely small amounts of clean labels are available. 
\end{itemize}

% \kai{significance testing..}
%We compare {\m} with several models, which focus on learning \textit{discriminative feature representations} from the message texts that are fed into a logistic regression classifier to predict the intent. 

We use following encoders for learning content representations:
\begin{itemize}
    \item {AvgEmb}:  AvgEmb learns the representation with the average word embedding of all the words in the message text. This is also the base model used in GLC~\cite{hendrycks2018using}.
    \item {BiLSTM}~\cite{graves2005framewise}: BiLSTM is a bi-directional recurrent neural network that learns the long-term dependencies in the text comprising of a sequence of tokens. During encoding, the input to the BiLSTM are the word embedding vectors of the input text.  
    % \item \textbf{GLC-AvgEMb}~\cite{hendrycks2018using}: GLC stands for Gold Loss Correction, which estimates a label corruption matrix to model the correlation between weak labels and clean labels, and forward a label correction layer to predict the unseen true labels. 
    % \item \textbf{DS-AvgEMb}:
    % \item \textbf{GLC-BiLSTM}: GLC-BiLSTM is a variant of GLC, with the prediction model as a BiLSTM network.
    % \item \textbf{DS-BiLSTM}: DS-BiLSTM is a bi-directional LSTM model for multi-task learning. The tasks include the weak label prediction and golden label prediction, and the representation learning layers are shared.
\end{itemize}
We truncate or pad the text to have a uniform length of 128 tokens. For both AvgEmb and BiSLTM, we use the pre-trained 300-dimensional GloVe~\cite{pennington2014glove} embeddings to initialize the embedding matrix, and fine-tune it during training. For BiLSTM, we use one hidden layer and set the number of the hidden states as 300, and take the last hidden state as the final feature representation. We also employ a fully connected layer with 300 hidden states after the BiLSTM network to capture the interaction across feature dimensions. We use cross-entropy loss as $\mathcal{L}$  in all settings. 
For self-paced learning, we set the number of epochs to $10$ for each run with a specific $\lambda$ varying from $[0.1, 0.2, \cdots]$. We vary hyper-parameter $\alpha\in\{0.1,1,10\}$ and choose the one that achieves the best performance on the validation set.

\subsubsection{Baselines:} We compare {\m} against the following baselines. 
The first set of baselines we consider is using the same base model as in {\m}: a three-layer neural network with word embeddings, an encoder (AvgEmb or BiLSTM) and a softmax layer for classification. We use this model in the following settings:
%For each of these models, we consider the following settings: (1) training with manually annotated \textbf{clean} data only; (2) training with \textbf{weakly} labeled data using labeling functions on a large pool of unlabeled instances only; and (3) training with both the \textbf{clean} and \textbf{weakly} labeled data. Several works have utilized these methods to leverage weak~\cite{10.1007/978-3-319-68155-9_16, mitra2017learning} or out-of-domain~\cite{daume2009frustratingly, hassan2013personalized} labeled data. Additionally, we use the Golden Loss Correction \textit{GLC} method~\cite{hendrycks2018using} as a baseline. Finally, we evaluate two variants of the proposed model: {\m-Base} and {\m}. We describe each model included in our experiments in detail here:

\begin{itemize}
    \item {Clean}: Model trained on only the clean instances.
    \item {Weak}: Model trained on only the weak labels derived from user interactions. Weak labels are treated as regular clean labels.
    \item {Clean+Weak}: In this setting, we simply merge both the sets (essentially treating the weak labels to be as reliable as the clean ones) and use them together for training.
    \item {Pre-Weak}: We first pre-train the model on the weakly labeled instances. Then we take the trained model and fine-tune all the parameters in all the layers end-to-end on the clean instances.
    \item {IWT}: In Instance-Weighted Training (IWT), we assign sample weights to each of the instances during learning. For this, we modify Equation~\ref{eq:0} as follows\footnote{Results are reported with $u(x)=10 \forall x$ and $v(\Tilde{x})=1 \forall \Tilde{x}$ with minimum error on the validation set.}.\\
        $\min_{\theta, \gamma_c, \gamma_w} \mathbb{E}_{(x,y)\in\mathcal{D}}[u(x)\cdot\mathcal{L}(y,f_c(enc(\mathbf{x})))]+\alpha\mathbb{E}_{(\mathbf{\Tilde{x}},\Tilde{y})\in\mathcal{\Tilde{D}}}[v(\Tilde{x}) \cdot \mathcal{L}(\Tilde{y},f_w(enc(\mathbf{\Tilde{x}})))]$\\
    with $u > v$ forcing the model to focus more on the clean instances during learning. Note that the Clean+Weak baseline is a special case with $u(x)=1 \forall x$. %\kai{clarify $\alpha$. Is this $\alpha$ the same with the $\alpha$ in Eqn 4? Should we mention what is the $\alpha$ we finally chose in stead of saying $\alpha>1$?}.
    \end{itemize}
    
The next baseline is the Gold Loss Correction (GLC)~\cite{hendrycks2018using} that estimates a label corruption matrix to model the correlation between weak and clean labels, which can be used to predict the unseen true labels.
    %\item {\m-Base}: It jointly leverages the weak and clean labels via two different loss functions to model their individual uncertainties and a shared representation (encoder) layer to model inter-dependencies.
Finally, we report results from our full model {\m}.
%Our final model {\m} extends {\m-Base} by refining the weak labels using GLC and training the network end-to-end leveraging both clean and weak labels. 
    % \item \textbf{{\m}}: 
% \end{itemize}
% \ms{Are we also going to compare our numbers with previously published numbers on these datasets too?}\kai{the most recent one is the SIGIR paper, which considers a sentence-level intent, while we consider a message-level intent. Thus, it's not comparable. We explain this here}

%It is worth mentioning that we cannot directly compare the recent model on email intent detection proposed in~\cite{wang2019context}, which considers the sentence-level intent prediction. Since our weak supervision is meaningful as the message-level signal, so we compare the baselines for message-level intent prediction.
%sm: The dataset in the above paper is not public. We should reduce references to the work and setup to preserve double blindness of the paper.
%sm: + the above models could be used for our use-case also as baselines for sentence-level tasks.

\subsection{Impact of Weak Supervision}

\begin{figure*}[htbp!]
\vspace{-0.15cm}
	\centering
	\subfigure[RI]{
	\includegraphics[width=0.28\textwidth]{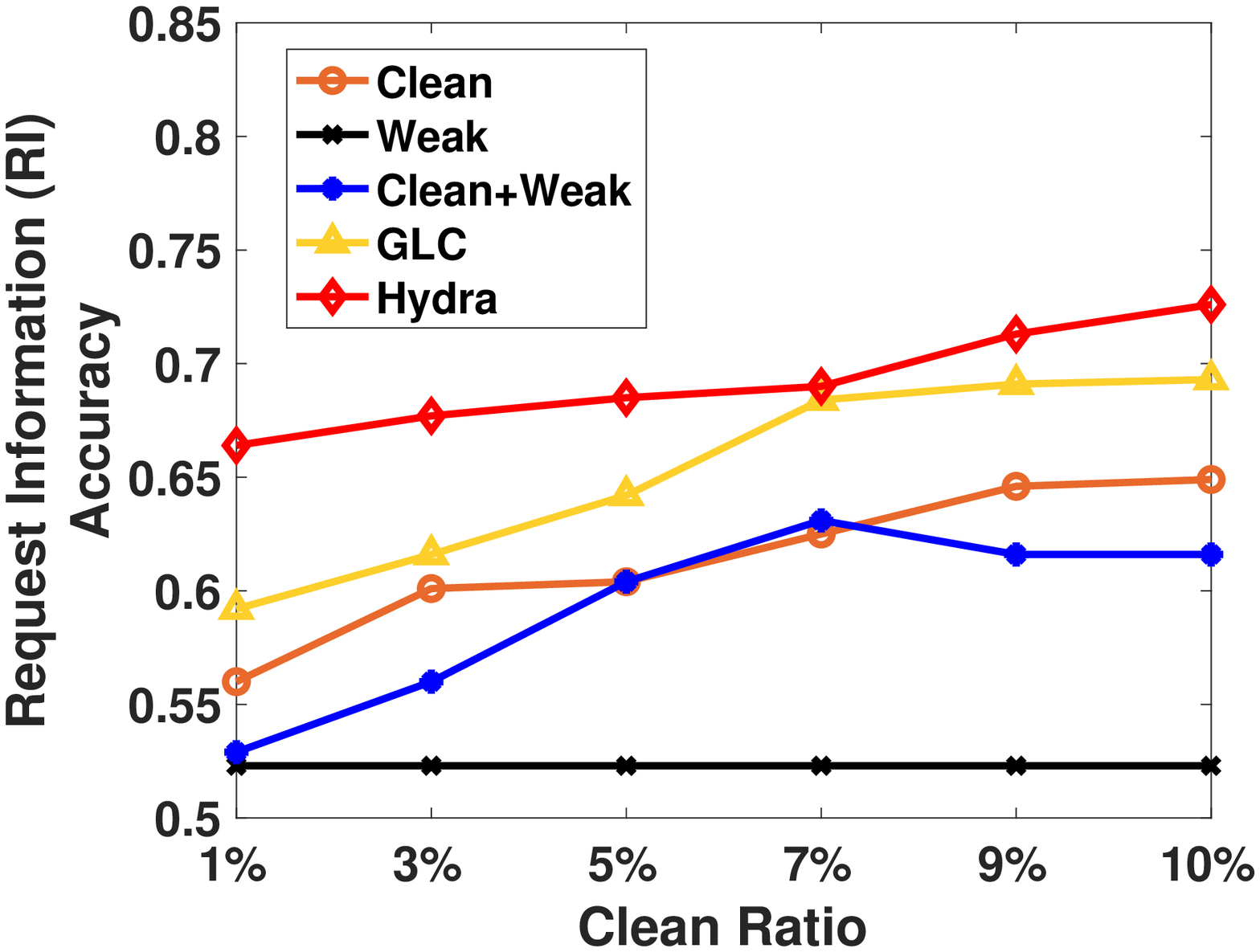}}
	%  	\vspace{-0.35cm}
		\subfigure[SM]{
	\includegraphics[width=0.28\textwidth]{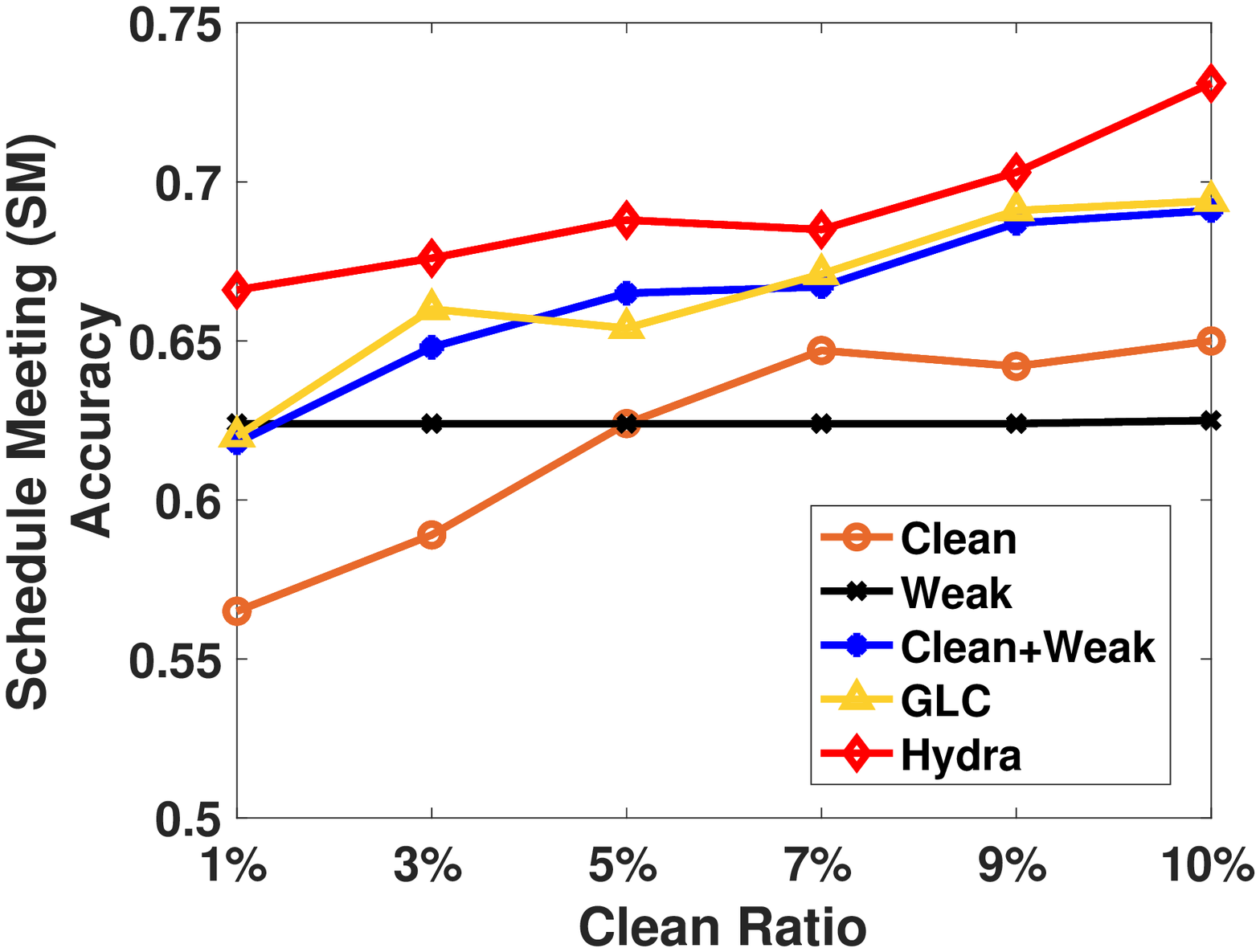}}
	%  	\vspace{-0.35cm}
		\subfigure[PA]{
	\includegraphics[width=0.28\textwidth]{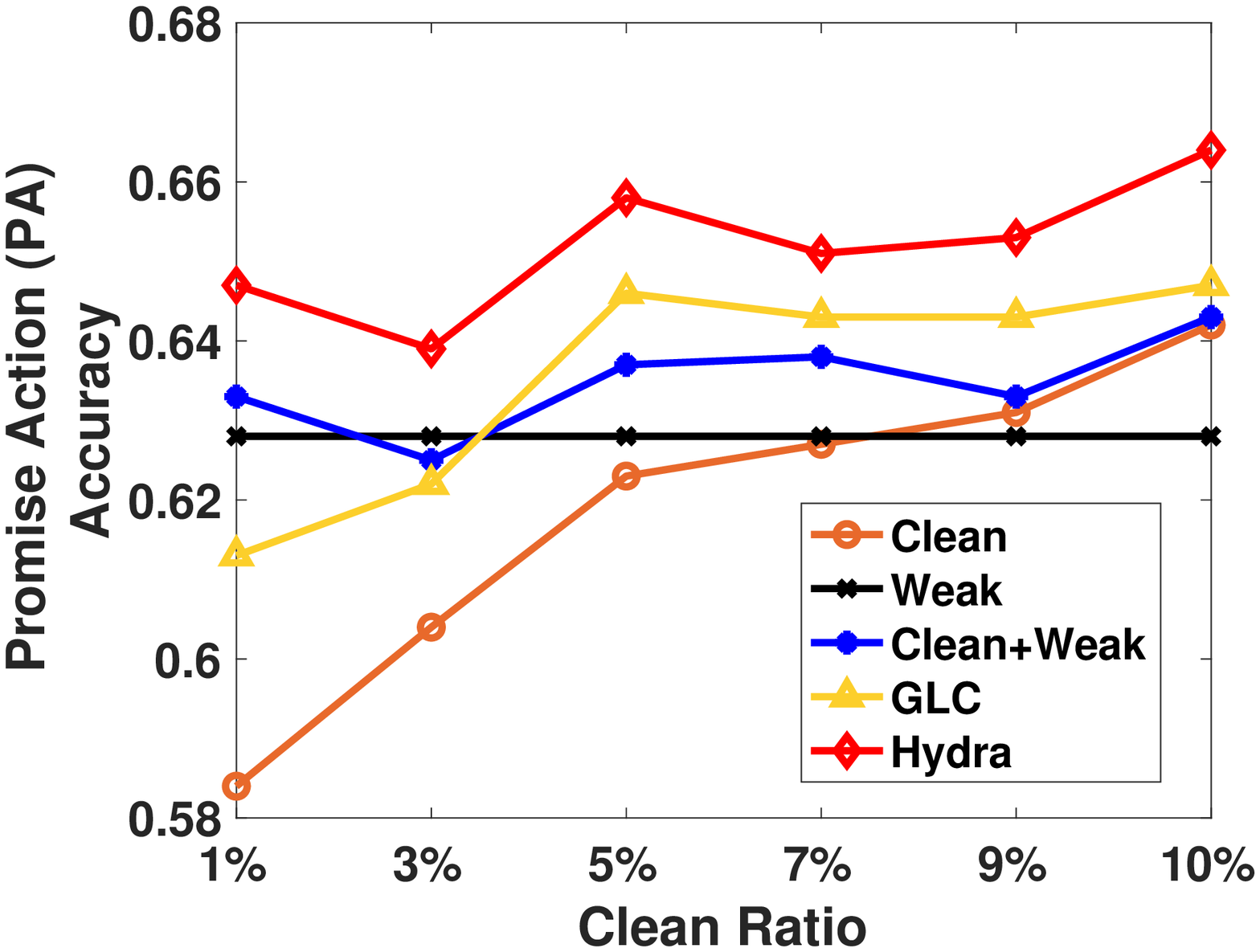}}
% 		\subfigure[Encoder=BiLSTM on RI]{
% 	\includegraphics[width=0.28\textwidth]{ri_bilstm_new.eps}}
% 			\subfigure[Encoder=BiLSTM on SM ]{
% 	\includegraphics[width=0.28\textwidth]{sm_bilstm_new.eps}}
% 		\subfigure[Encoder=BiLSTM on PA ]{
% 	\includegraphics[width=0.28\textwidth]{pa_bilstm_new.eps}}
 	\vspace{-0.25cm}
	\caption{Classification results with varying clean ratio for different tasks (enc=AvgEmb) (best viewed in color). We keep the number of weakly labeled instances fixed, and vary the amount of cleanly labeled ones for any given setting.}\label{fig:golden_ratio}
	\vspace{-0.2cm}
\end{figure*}%\todo{SM: Put Encoder=BiLSTM/AvgEmb. Change the legends as Hydra and Hydra-Base.}

%To answer \textbf{EQ1}, we compare the proposed framework {\m} with the representative methods introduced in Section~\ref{sec:setting} for intent classification.  For this experiment, we use all the available clean and weakly labeled instances for each setting. %We take all the clean data and weak supervision data to train all the baselines.

Table~\ref{tab:performance} shows the performance comparing {\m} to other models in different settings. From therein we make the following observations based on varying clean ratio. 
% \kai{may discuss the different setting with SIGIR paper here briefly}
\begin{itemize}
    \item Training only on the clean samples (even though they are much smaller in size) achieves better performance than training only on the weakly labeled ones on an aggregate across all the tasks and settings (demonstrated by Clean $>$ Weak).% hold for both base models including AvgEMb and BiLSTM, in RI, SM, and PA tasks.
    \item Incorporating weakly labeled data even by simple aggregation with clean data like (\textit{Clean + Weak}), pre-training (\textit{Pre-Weak}) and instance weighting (\textit{IWT}) improves model performance on an aggregate over that of using only the clean or weak data.
    \item More sophisticated methods of integrating weakly labeled data with clean data gradually improves the performance on an aggregate across all the tasks (demonstrated by {\m} $>$ GLC $>$ IWT $>$ Pre-Weak $>$ Clean+Weak $>$ Clean $>$ Weak). 
    \item Finally, irrespective of the clean ratio, {\m} achieves the best performance in all settings. 
\end{itemize}

%However, with only few clean instances (e.g., clean ratio = 0.01), the models using simple combination of weak labels (e.g., Weak, Clean+Weak, Pre-Weak and IWT) perform better than using only the clean samples in many cases.

%We observe that {\m} $>$ \{GLC, IWT, Pre-Weak, Clean+Weak\} for all the encoders across all the tasks.

%\vspace{-0.2cm}
\subsection{Impact of Clean Data Ratio}

In real world scenarios, we often have a limited amount of annotated data, that would vary depending on the task and the domain, and a large amount of unlabeled data. In this experiment, we explore how the performance of {\m} changes with varying amount of clean data. To this end, we fix the number of weakly labeled instances and change the number of clean instances for each setting, thereby, varying the clean ratio between $[1\%,3\%,5\%,7\%,9\%,10\%]$.
%For {\m} the weak labels come from GLC with label correction. 
Figure~\ref{fig:golden_ratio} shows the performance of different models using AvgEmb as the encoder. The graphs for BiLSTM encoder are similar and were omitted for space considerations. Since \textit{Clean + Weak}, \textit{Pre-Weak} and \textit{IWT} have similar graphs, we only show results from the former. %\kai{We have similar observations when using BiLSTM as the encoder, so we only show the figures with AvgEmb as the encoder.}
% The corresponding clean ratio is 0.1 which achieves the best performance for GLC among all the golden ratios. \guoqing{Maybe remove this?} 
We make the following observations from Figure~\ref{fig:golden_ratio}:
\begin{itemize}
\item With increasing clean ratio, the performance increases for all models (except \textit{Weak} which uses a fixed amount of weakly labeled data). This is obvious as clean labels are more reliable than the weak ones.
\item {\m} consistently outperforms all other methods accessing both clean and weak instances.
%achieves the best performance compared with all other baselines using both clean and weakly labeled data, i.e., %{\m} $>$ {\m}-Base, {\m} $>$ GLC, and {\m} $>$ Clean+Weak. This shows that {\m} can more effectively utilize the clean and weak labels via its multi-source learning framework.
%\item We observe that methods using Clean+Weak where we treat the weak labels to be as reliable as clean ones may not necessarily perform better than only $Clean$ or only $Weak$. 
% \ms{Clean+Weak vs. Clean and weak could be confusing for the reader.} 
\item Simple aggregation of clean and weak sources of supervision (e.g, Clean+Weak) without accounting for the source's uncertainties (as modeled separately in {\m}) may not improve the prediction performance. %The graphs for $PreWeak$ and $IWT$ are similar to that of Clean+Weak and are omitted. 
\item The performance gap between using only the clean labels and {\m} using both clean and weak labels decreases with the increase in clean ratio. Note that {\m} is more effective when the clean ratio is small which is a more realistic setting with a small amount of labeled data.
\end{itemize}

\subsection{Ablation Analysis}

\noindent {\bf Self-paced Learning.} In order to understand the contribution of self-paced learning in the {\m} framework, we perform another experiment. In this we train {\m} leveraging all the clean and weak labels jointly without any curriculum. At each epoch, we sample batches with equal number of instances from clean and weakly labeled data and train {\m} end-to-end optimizing Equation~\ref{eqn:obj}.
%Label correction and self-paced learning both contribute to the effectiveness and robustness of {\m}. In order to understand their contribution, we perform an ablation test removing one component at a time and report the results in Table~\ref{tab:ablation}. Here we report aggregated results over all the settings with different encoders and intents for two clean ratio settings.
\begin{table}[t!]
    \centering
    \caption{Ablation analysis for {\m}. First row in each section shows {\m} with self-paced learning and GLC. Results are average across all tasks \& encoders for a given clean ratio.}
    \begin{tabular}{clc}
    \toprule
        Clean Ratio (Setting)& Components & Accuracy  \\\midrule
        \multirow{3}{*}{\bf{10\% (All)}} & {\m} & {\bf 0.716}\\
         
          & \ \   $-$ self-paced learning & 0.690 \\
          & \ \   $-$  GLC & 0.688 \\
         
         \midrule
         \multirow{3}{*}{\bf{1\% (Tiny)}} & {\m}  & {\bf 0.643} \\
         
         &  \ \   $-$ self-paced learning & 0.631\\
        
         & \ \   $-$  GLC  & {0.632}\\ 
         \bottomrule
    \end{tabular}
    \label{tab:ablation}
    \vspace{-0.25cm}
\end{table}
From Table~\ref{tab:ablation}, we observe self-paced learning to improve the performance of {\m} on aggregate across different tasks for different values of clean ratio. We observe the self-paced learning to perform much better at higher values of clean ratio that contributes a larger set of clean samples for training. This results from our training schedule (refer to Section~\ref{sec:training}) where we initially train {\m} for a few epochs on the clean data to trace the initial model space for $w$. 

\noindent{\bf Gold Loss Correction (GLC).} In this, we remove the GLC component from {\m} and report the performance in Table~\ref{tab:ablation}. Similar to self-paced learning, we observe a similar performance loss at different values of the clean ratio. Note that both GLC and self-paced learning capture the noise in the weakly labeled instances to inform {\m} during training. However, the GLC component is learnt offline where {\m} uses only a fixed copy of the label corruption matrix. Whereas, self-paced learning updates all the parameters in {\m} during training.

\subsection{Additional Experiments}
%In this section, we analyze some important parameters and their effects on the performance of {\m}.

\noindent\textbf{Relative importance of clean and weak sources:} The hyper-parameter $\alpha$ controls the relative importance of the losses computed over the labels from the clean and weak sources. We observe that a larger value of $\alpha$ results in a better performance when the clean ratio is small, and vice versa when the clean ratio is large. This shows that {\m} relies more on the weak labels when the amount of clean labels is less. This dependency decreases with greater availability of reliable annotations. For example, when the clean ratio is 1\% and 3\% (see Figure~\ref{fig:golden_ratio}), {\m} achieves the best performance with $\alpha=10$; whereas, with a clean ratio ranging from 5\% to 10\%, $\alpha=1$ leads to the best performance consistently across all the tasks.  

\noindent\textbf{Impact of the amount of weak labels:} 
In Figure~\ref{fig:golden_ratio}, we fixed the amount of weak labels and varied the amount of clean labels for different tasks depicting performance improvement with increase in the amount of reliable labels. In order to assess the impact of the amount of available weak labels on the performance of {\m}, 
% we look at the relative performance improvement for different tasks with varying amount of weak labels. From Tables~\ref{tab:data} and~\ref{tab:performance}, we observe that {\m} achieves a relative performance improvement of $12\%$ over using only clean labels in the RI task where the amount of available weak labels is five times the amount of clean labels; whereas, for the PA task the corresponding performance improvement is only $2.6\%$ given lesser amount of weak labels than the clean ones. 
% The above discussion, however, ignores the relative difficulty of the different tasks.  Therefore, 
we perform another experiment where we fix the task and the amount of clean labels, and vary the amount of weakly labeled instances. We perform this experiment with {\m} and BiLSTM encoder for the RI task. From Table~\ref{tab:ri_weak}, we observe that the performance of {\m} improves with increase in the amount of weak data, further demonstrating the importance of weak supervision.

\begin{table}[t!]
\centering \caption{{\m} (enc = BiLSTM) on RI task with fixed amount of clean data (i.e. all the 1800 clean instances) and varying percentage of weak labels.}
\label{tab:ri_weak}
\begin{tabular}{c|ccccccc}
\toprule
\% weak labels & 0\% & 20\% & 40\% & 60\% & 80\% & 100\% \\
\midrule
\# weak labels & 0 & 3,240 & 6,480 & 9,720 & 12,960 & 16,200 \\
\midrule
Accuracy & 0.688 & 0.711 & 0.732 & 0.744 & 0.744 & 0.804\\
\bottomrule
\end{tabular}
\vspace{-0.22cm}
\end{table}
% alpha
% relative gain
% case study: examples 
% significance testing

% \begin{figure}[htbp!]
% 	\centering
% 	\subfigure[AvgEMb]{
% 	\includegraphics[width=0.35\textwidth]{avgemb_relative.png}}
% % 	\hspace{-0.3cm}
% 	\subfigure[BiLSTM]{
% 	\includegraphics[width=0.35\textwidth]{bilstm_relative.png}}
% 	\caption{The relative performance gain of {\m} and GLC to that of only Clean on RI, SM, and PA tasks.}\label{fig:relative}
% \end{figure}
% \aha{this seems redundant given Table 5 and the formatting/resolution needs to improve}
% \begin{figure}
%     \centering
%     \includegraphics{}
%     \caption{Caption}
%     \label{fig:my_label}
% \end{figure}{}

% The weight that controls the relative importance of clean and weak labels
% parameter analysis on alpha

\subsection{Domain Transfer}
\label{subsec:dom_gen}

Now we want to test the transferability of {\m}. In other words, we want to test if the weak data from one domain can help in intent prediction in another domain.
% if the model trained on clean and weak signals in one domain can be applied to another.
To this end, we train {\m} using the clean data in Enron and weak data in Avocado,
% in Avocado with different clean ratio settings, 
and test the corresponding model on Enron. As baseline, we consider the base model used in {\m} trained using only the clean labeled data in Enron (see baseline description in Section~\ref{sec:setting} and data statistics in Table~\ref{tab:data}). All models have the same test set of 908 instances in Enron.
Table~\ref{tab:domain_gene} shows results for the task of scheduling meeting intent with different clean data ratio. We observe that {\m} trained on clean data from Enron and weak data from Avocado, and tested on Enron shows better results than that trained only on the clean data from Enron. This shows that (i) {\m} transfers well across domains, and (ii) weak signals from one domain can be leveraged to improve the performance of models in another domain transferred via shared encoders. We also perform an experiment, where we train {\m} on Avocado using all the available clean and weak instances, and test on Enron. In this zero-shot transfer setting (with no labeled training data requirement for Enron), {\m} obtains an accuracy of $0.738$. Comparing this setting to the above settings in Table~\ref{tab:domain_gene}, we observe that (i) {\m} performs better than En. $\rightarrow$ En. (using only clean data from Enron) ($0.717)$ demonstrating transferability, and (ii) worse than Av. $\rightarrow$ En. (using clean data from Enron and weak data from Avocado) ($0.821$) demonstrating the benefit of adding some target domain clean data.

%When we compare {\m} in this setting to the model trained and tested on Enron, we make similar observations as in Table~\ref{tab:domain_gene}, further demonstrating the generalizability of our framework.  

\begin{table}[t!]
% \small
\centering \caption{Domain transfer for {\m} (enc=BiLSTM) on SM task. Av. $\rightarrow$ En. denotes {\m} trained on clean data in Enron and weak data in Avocado, and tested on Enron. Whereas En. $\rightarrow$ En. denotes the model trained on only the clean data in Enron, and tested on Enron.} 
\vspace{-1em}
\label{tab:domain_gene}
\begin{tabular}{c|cc|cc}
\toprule
 & \multicolumn{2}{c}{En. $\rightarrow$ En.} & \multicolumn{2}{|c}{ Av. $\rightarrow$ En.} \\
\midrule
%  Clean Ratio & - & 1\% & 10\%(All)   \\
\# clean training labels & 82 & 908 & 82& 908\\
 %\midrule
\# weak training labels & 0& 0 & 8,176& 8,176\\
 \midrule
% \rowcolor{shadecolor}
%  Accuracy &0.533 &0.738 & 0.785 & 0.827 \\
Accuracy &0.714 &0.717 & 0.752 & 0.821 \\
\bottomrule
\end{tabular}
\vspace{-0.4cm}
\end{table}

% \kai{add the table for one task and discussion}

\section{Related Work} \label{sec:related} %\todo{Put this section in the end}
In this section, we briefly review the related work on email intent detection, weak supervision and learning from user interactions in other applications such as web search. %We cover each line of research in turn and situate our work with respect to previous work.

% \kai{add related work to additional 2/3 column}

%Varma \textit{et al.} propose to exploit structure learning on multiple-source weak supervision signals for improving prediction performances~\cite{varma2019learning}.

%In this paper, by incorporating the label correction approach, we develop the novel framework of {\m} for learning from trusted annotated data and weak labeled data simultaneously. 
%Our proposed framework {\m} can jointly learn the representations with clean data and corrected weak-labeled data for improving prediction performance.

% ~\cite{ratner2018training} model the relationships by treating weak labeling tasks as the sub-tasks of original task and utilize a hierarchical multi-task learning scheme.
% ~\cite{zamani2018neural} utilize the signals of unsupervised models as weak supervision for improving the performance of query performance prediction task.
% image domain
% text domain
%\aha{Given this is being submitted to WSDM, should we cover leveraging user interactions for training search systems/rankers?}
%\guoqing{A few citations added}
\subsection{Email Intent Classification}
Email understanding and intent classification has attracted increasing attention recently. Dabbish \textit{et al.} conduct a survey on 124 participants to characterize email usage~\cite{dabbish2005understanding}. They highlight four distinct uses of email intents like project management, information exchange, scheduling and planning, and social communication. Detecting user intents, especially action-item intents~\cite{bennett1972detecting}, can help service providers to enhance user experience. Recent research focuses on predicting actionable email intent from email contents~\cite{wang2019context,lin2018actionable}, and identify related user actions such as reply~\cite{yang2017characterizing}, deferral~\cite{sarrafzadeh2019characterizing}, re-finding~\cite{mackenzie2019exploring}. Wang \textit{et al.} model the contextual information in email text to identify sentence-level user intents. Lin \textit{et al.} build a reparameterized recurrent neural network to model cross-domain information and identify actionable email intents. In a more finer-grained level, Lampter \textit{et al.}~\cite{Lampert:2010:DEC:1857999.1858140} study the problem of detecting emails that contain intent of requesting information, and propose to segment email contents into different functional zones. More recently, Azarbonyad \textit{et al.}~\cite{azarbonyad2019domain} utilize domain adaptation for commitment detection in emails. They demonstrate superior performance using autoencoders to capture both feature- and sample-level adaptation across domains. In contrast to all these models trained on manually annotated clean labels, we develop a framework {\m} that leverages weak supervision signals from user interactions for intent classification in addition to a small amount of clean labels.

% As another type of communication, digital assistants (e.g., Cortana, Siri) allow people to express the intents and trigger the actions such as setting calendars, requesting meetings, etc~\cite{khabsa2018identifying}. Rishabh \textit{et al.} Propose to identify user sessions using a Gaussian mixture model and accurately predict the task boundaries~\cite{khabsa2018identifying}.  Recently, Ryen \textit{et al.} estimate the time duration~\cite{white2019taskduration}, and the completion~\cite{white2019task} of tasks in digital assistants, which go beyond understanding users' intents and automatically manage and optimize them.

%In this paper, we study the novel problem learning with weak supervision from user interactions for email intent detection. Different from most existing work that extract features from the email content, we explore the additional weak-supervision signals from user interactions for improving the performance of intent detection.

\subsection{Learning with Weak Supervision}
% Supervised learning techniques have achieved great success when there is strong supervision information like large amount of training examples with ground-truth labels.  In real tasks, however, collecting supervision information requires costs, and thus, it is usually desired to be able to do weakly supervised learning. 
Most machine learning models rely on the scale of labeled data to achieve good performance where the presence of label noise~\cite{nettleton2010study} or adversarial noise~\cite{reed2014training} can cause a dramatic performance drop. Therefore, learning with noisy labels has been of great interest to the research community for various tasks~\cite{natarajan2013learning,frenay2013classification,meng2019weakly}. Some of the existing works attempt to rectify the weak labels by incorporating a loss correction mechanism~\cite{sukhbaatar2014training,patrini2017making}. Sukhbaatar \textit{et al.}~\cite{sukhbaatar2014training} introduce a linear layer to adjust the loss and estimate label corruption with access to the true labels~\cite{sukhbaatar2014training}. Patrini \textit{et al.}~\cite{patrini2017making} utilize the loss correction mechanism to estimate a label corruption matrix without making use of clean labels. Other works consider the scenario where a small set of clean labels are available~\cite{li2017learning,charikar2017learning,hendrycks2018using,ren2018learning}. For example, Veit \textit{et al.} use human-verified labels and train a label cleaning network in a multi-label classification setting. Recent works also consider the scenario where weak signals are available from multiple sources~\cite{ratner2017snorkel,varma2019learning,ratner2018training} to exploit the redundancy as well as the consistency in the labeling information. %In contrast, the weak signals in our work are derived from user interactions. In addition, our work does not make any strong assumptions about the structure of the noise or depend on the availability of multiple weak sources to model corroboration. %User interaction has been widely used to build and improve Web search and recommender systems, including using user clicks to improve data quality for learning to rank~\cite{xu2010improving}, predicting and improving search results~\cite{agichtein2006learning,agichtein2006improving}, and incorporating user interactions for better recommendation ~\cite{rendle2010pairwise,jiang2012social}.
We build on top of previous work in the area of learning from weak supervision. More specifically, we focus on an application (email intent detection) where weak supervision can be obtained from user interaction signals.
Second, we focus on a setup where a small amount of clean labeled data is available and propose methods to combine it with larger datasets with weak labels to improve the overall performance. In addition, our work does not make any strong assumptions about the structure of the noise or depend on the availability of multiple weak sources to model corroboration.

\subsection{Learning from User Interactions in Web Search}

Modern web search engines have heavily exploited user interaction logs to train and improve search systems~\cite{agichtein2006improving,joachims2007evaluating}. In fact, previous work  ~\cite{hassan2010beyond} showed that models using click behaviors are more predictive of goal success than using document relevance. Motivated by the success of deep learning methods, several studies have focused on developing deep ranking models for webs search. Since such models require access to large amounts of training data, they opted for using click data ~\cite{10.1007/978-3-319-68155-9_16,mitra2017learning}  or the output of an unsupervised ranking model, BM25, as a weak labels~\cite{dehghani2017neural}.

Our work is similar to this line of work in that they both try to leverage user interaction data to improve a machine learning system. They are also different in many ways. First, we focus on intent classification in email text as opposed to ranking. Additionally, we focus on methods that combine clean-labeled data and weak-labeled data as opposed to just using implicit feedback data like clicks. Finally while clicks were shown to be more accurate than other types of user interaction signals, they suffer from several types of biases (e.g. snippet as, positing bias, etc.). Understanding the difference between such user interaction signals and clicks is an interesting direction for future research.

\section{Conclusions and Future Work}\label{sec:conclusion}
In this paper, we leverage weak supervision signals from user interactions to improve intent detection for emails. We develop an end-to-end robust neural network model {\m} to jointly learn from a small amount of clean labels and a large amount of weakly labeled instances derived from user interactions. Extensive experiments on a real-world email dataset, Avocado, show {\m} to not only outperform state-of-the-art baselines but also its effectiveness in transferring the weak signals to another domain, namely Enron. %Although our experiments are focused on email communication, our techniques can also be applied to similar settings involving weak and clean labeled data.

There are several directions for further investigation. First, we can extend our framework to multi-task learning where all of the above intent classification tasks can be learned jointly along with multiple sources of weak supervision. It is also worth exploring combining label correction and multi-source learning jointly instead of a two-stage approach. Second, understanding the nature of different sources of weak supervision is valuable for learning in different application domains. For example, in web search, user clicks can be a relatively accurate source of weak supervision, but may suffer from presentation bias; while for email data, user interactions are less accurate but may not suffer from the same biases.
\balance
\bibliographystyle{plain}
\bibliography{ref}

\end{document}